\documentclass[acmlarge]{acmart}

\AtBeginDocument{%
  \providecommand\BibTeX{{%
    \normalfont B\kern-0.5em{\scshape i\kern-0.25em b}\kern-0.8em\TeX}}}

\acmJournal{JOCCH}
\acmVolume{14}
\acmNumber{1}
\acmArticle{2}
\acmMonth{2}

\usepackage{soul}
\usepackage{amsmath}

\begin{document}
\settopmatter{printacmref=false}
\renewcommand\footnotetextcopyrightpermission[1]{}
\pagestyle{plain}

\title{Event-based Access to Historical Italian War Memoirs}


\author{Marco Rovera}
\email{rovera@hdm-stuttgart.de}
\affiliation{
    \institution{Department of Computer Science, University of Torino}
    \streetaddress{Corso Svizzera, 185}
    \city{Torino}
    \country{Italy}
    \postcode{10149}
}
\affiliation{%
  \institution{Institute for Applied AI (IAAI), Stuttgart Media University}
  \streetaddress{Nobelstrasse, 10}
  \city{Stuttgart}
  \country{Germany}
  \postcode{70569}
}

\author{Federico Nanni}
\affiliation{
  \institution{Data and Web Science Group (DWS), University of Mannheim}
  \streetaddress{B6, 26}
  \city{Mannheim}
  \country{Germany}}
\email{federico@informatik.uni-mannheim.de}

\author{Simone Paolo Ponzetto}
\affiliation{%
  \institution{Data and Web Science Group (DWS), University of Mannheim}
  \streetaddress{B6, 26}
  \city{Mannheim}
  \country{Germany}}
\email{simone@informatik.uni-mannheim.de}

\renewcommand{\shortauthors}{Rovera et al.}

\begin{abstract}
The progressive digitization of historical archives provides new, often domain specific, textual resources that report on facts and events which have happened in the past; among these, memoirs are a very common type of primary source. In this paper, we present an approach for extracting information from Italian historical war memoirs and turning it into structured knowledge. This is based on the semantic notions of events, participants and roles. We evaluate quantitatively each of the key-steps of our approach and provide a graph-based representation of the extracted knowledge, which allows to move between a Close and a Distant Reading of the collection.  
\end{abstract}



\maketitle

\section{Introduction}
The growing interest of cultural institutions in the digitization of archival documents and resources, along with the availability of new, born-digital textual materials about historical topics, raises the question of how to provide the users with an account of the knowledge contained in such collections. From a computational point of view, one of the challenges lies in the capability of developing models and techniques to automatically extract information from historical (most of the time OCR-digitized) documents and turn such information into knowledge that can be easily accessed, queried and visualized. 

The possibility to deal with thousands of books or documents
at once \cite{crane2006you} was the main foreground for the idea of ``Distant Reading'' \cite{moretti2003graphs,moretti2013distant}. The Distant Reading approach suggested a new paradigm based on the application of computational tools for analyzing large literary collections, fostering a new approach for scholars in the humanities methodologically closer to the social sciences.

During the last two decades, with the growing availability of applications following the general framework of Distant Reading in historical research \cite{moretti2013distant,graham2015exploring}, several examples of advanced access to digitized collections through the use of text mining technologies have been presented (see for instance the works by \citet{wilkens2013geographic,blevins2014space,kaufman2015everything}). 
Though results have been fascinating, the Distant Reading paradigm has been also strongly criticized, especially for its bias towards quantification and scientification of humanities’ research problems \cite{posner2015humanities}. Moreover, while nowadays established Distant Reading approaches such as Topic Modeling and Named Entity Recognition support the users in going beyond traditional keyword searches, often these text mining techniques produce only coarse-grained macro-overviews of the information contained in the collection under study, for instance by providing a list of the most frequently mentioned entities or the most recurrent topics (as already discussed by \citet{janicke2015close,nanni2016semi}). Additionally, these methods generally lack the possibility of rendering back semantic information in a more fine-grained way (e.g., by retrieving highly relevant sentences). The possibility of moving between (macro-) quantitative analysis and (micro-) qualitative studies is now a relevant topic both in Digital Humanities and Computational Linguistics, especially because it could allow to get a better understanding of the model, the corpus and the underlying assumptions.

With this in mind, the goal of our work is to propose an event-based Distant Reading approach that additionally allows to obtain a closer, semantic and content-aware reading of the sources behind the creation of each specific pattern in the output. This, in order to \emph{a)} overcome the previously mentioned criticism that Distant Reading approaches keep the reader too far away from the source \cite{spanos2017against}, \emph{b)} offer a better understanding of how our method works and \emph{c)} recognize the value of both quantitative and qualitative types of approaches to historical sources \cite{janicke2015close}. In section 7, we discuss the advantages of such representation in a series of scenarios that allow both Close and Distant Reading of the corpus. The long-term aim of this effort is to support scholars and other stakeholders in Digital and Public History to dive into large textual collections and to fulfill their information needs by exploring many different textual sources in a consistent way.
To do so, we develop a pipeline for extracting events and event participants, along with their roles, from text; additionally, we resolve Named Entity co-reference in order to be able to match each unique entity to the set of events it participates in. Finally, we test this pipeline in a real-world scenario on a corpus of Italian war memoirs concerning the Second World War, digitized for this purpose. We present an extensive evaluation and error analysis of each step of our pipeline.\smallskip

\noindent \textbf{Outline.} The rest of the paper is structured as follows: in Section 2 we discuss some of the most relevant previous works related to our study from the fields of Text Mining and Event Extraction. Section 3 describes the textual corpus and the knowledge resources used in our system. Section 4 presents the way we model events in our work, which led to the creation of a lexical resource for extraction of events and participants. In Section 5 the extraction pipeline is sketched in all its components and in Section 6 evaluation results are discussed for each of its key-steps. In Section 7, we discuss some practical use case scenarios for event-based access to the extracted knowledge base; this allows moving from the textual collection to a network representing events, participants and roles, back again to the text. Section 8 concludes the paper and briefly discusses the reusability of our solution and resources by other researchers, along with the future directions we envisage for our work. Finally, Appendix A contains the list of digitized historical texts we employ in this work.
The gazetteers (4.2), the event-predicate dictionary (4.2), the final event data (5.3) and a Gephi event-entity graph (7), together with code to process them are available
on github\footnote{github.com/marcorovera/ita-resistance}. 

\section{Related Works}
In this section, we first offer an overview of related works on advanced access to digital library collections through the use of text mining methods. Next, we cover previous research on Information Extraction and Semantic Role Labeling, relevant for our work.\smallskip

\noindent \textbf{Advanced Access to Textual Collections.} During the last fifteen years Latent Dirichlet Allocation (LDA) topic modeling \cite{blei2003latent,newman2010evaluating} has arguably been the most popular text mining technique for corpus exploration in Digital Humanities (DH) and Digital Libraries (DL). This approach has been adopted in order to offer advanced access to scientific collections \cite{mann2006bibliometric}, proceedings of political debates \cite{greene2017exploring} and historical corpora \cite{nelson2010mining}. To face the limitations of LDA topics \cite{chang2009reading} and in order to extract information easier to interpret for final users, these communities have seen, in recent years, a growth in the combination of Named Entity Recognition (and whenever possible linking) with network analysis techniques. \citet{kaufman2015everything} relies on these approaches for examining the Digital National Security Archive (DNSA) Kissinger Collections, while \citet{menini2017ramble} use them for tracing the movements of popular historical figures mentioned in Wikipedia and \citet{ardanuy2014structure} for clustering novels by genres and authors.

Following the potential of such combination of Named Entity Recognition and network analysis technologies for offering advanced access to historical collections, in this paper we intend to go a few steps deeper into the semantic information that could be extracted from textual data. To do so, we identify and disambiguate mentions of entities with a domain-specific knowledge resource, tag them with the specific semantic role (depending on the event in which they appear) and highlight their network of relations.\smallskip

\begin{figure*}[h!]

\includegraphics[width=0.9\textwidth]{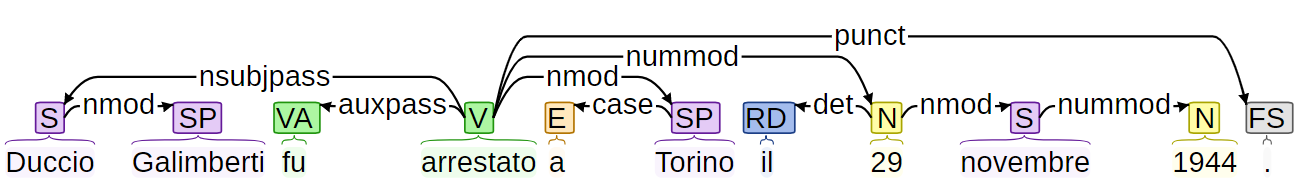}
\caption{Example of the correspondence between syntactic arguments of the verb and participants of the event denoted by the verb (translation: \textit{Duccio Galimberti was arrested in Torino on the 29th November 1944}). This image has been created using the Simpatico project demo, available at \url{http://simpatico.fbk.eu/demo/}.}
\end{figure*}

\noindent \textbf{Event Extraction.}
This paper focuses on the extraction of fine-grained events for providing advanced access to historical collections. The choice of events as a pivotal notion is motivated by the consideration that events are a “natural” structuring concept, at both a representation and a linguistic level, as they tie together time, space and participants; this observation holds true especially when dealing with historical texts. Events have been the focus of many different research areas, notably Computational Ontology, Computational Linguistics and Natural Language Processing, among others. Due to the magnitude of this research, we restrict our focus on recent interest from the Digital Library community in event-based collection building \cite{nanni2017building,nanni2018toward,nwala2017local,kanan2015big,gossen2017extracting} and its applications. In NLP and related areas, the notion of event has been modeled and applied to different subtasks, such as Question Answering \citep{yang2003structured}, Topic Detection and Tracking \citep{allan1998line, kumaran2005using}, Narrative Chain Induction \citep{chambers2008unsupervised, chambers2009unsupervised}, Entity Disambiguation \citep{roberts2010toponym}, Information Extraction and Retrieval \citep{spitz2016terms, spitz2017evelin}. Applications in this area are well exemplified by a research thread focusing on different aspects of events in historical texts and newswires, for example Extraction \citep{segers2011hacking, cybulska2011historical} and Coreference Resolution \citep{cybulska2013semantic, cybulska2015bag}. This thread was also part of NewsReader \citep{vossen2016newsreader}, a multilingual, EU-funded research project focused on Event and Information Extraction from newswires. 
Our effort, especially the design choice of focusing on fine-grained representations of events, is further encouraged by the findings provided in their work by Sprugnoli and Tonelli \cite{sprugnoli2017one}; in a multidisciplinary setting, the authors investigate the gap between computational techniques and the community of historians. They directly involve scholars from this area by conducting a survey, aimed at understanding how events are envisioned by the research community. One of the findings reported by the authors is that the fine-grained type of an event is considered important by historians, but existing annotation schemas (like ACE \cite{doddington2004automatic}, Event Nugget \cite{mitamura2015event} or TimeML \cite{pustejovsky2005specification,sauri2009annotating}) mostly provide broad event classifications. Building upon this previous study, our work aims at providing a real-world application of entity and event detection techniques on a newly digitized historical collection of war memoirs.

\noindent \textbf{Semantic Role Labeling and Frame Semantics.} Semantic Role Labeling (SRL) is the task of automatically assigning a semantic role to a portion of text in a sentence. A \textit{semantic role} is a label describing the thematic role played by a word or a group of words with respect to the main action or state described in the sentence; for example an \textsc{Agent} carries out an action, a \textsc{Patient} is subjected to an action or to its consequences, a \textsc{Source} marks the starting point of a movement. Since it requires a pre-defined set of semantic roles, along with a linguistic knowledge base in which relationships between lexical units (i.e. words) and semantic roles are formally made explicit, SRL is generally approached as a supervised task. Typical lexical knowledge bases for SRL in English are FrameNet \citep{baker1998berkeley}, VerbNet \citep{schuler2005verbnet} and Propbank \citep{palmer2005proposition}. Since such knowledge bases exist only partially for Italian, in this work we created a lightweight resource, focusing on three high-level event categories: movements, conflict events, and social events concerning membership and organizations. 

\section{Texts and Resources}

In this section we provide an overview of the collections and the knowledge resources employed in our work.

\subsection{Textual Collections}

The textual collection employed in this work is divided into three subcorpora, as presented below. All textual resources are in Italian. The central resource comprises historical war memoirs of Italian partisans from World War II in North-Western Italy (Memoirs and Memoirs-test). This is accompanied by two other related resources: biographic records (Biographies) of the participants to these events and encyclopedic entries on the topic (Wiki-Articles).
The Memoirs subcorpus is the one of interest for the scope of this work and the one which the system will be tested directly on. We consider the availability of such textual corpus of great importance for many reasons. In fact, we claim that one of the responsibilities of Digital History is to preserve and make accessible the past from multiple perspectives, by paying attention both to the individual and collective stories of women and men who acted in the past and to their relationship and role within the great events which ``make the history''. Moreover, our texts are one example of textual resources that are typically provided by historical archives and, as non-digital-born resources, they challenge current computational techniques which are often based on and oriented to digital sources (e.g. Wikipedia, Twitter) or standard texts (e.g. newspapers). The Biographies and the Wiki-Articles subcorpora have been employed for expanding the lexical resource for Event Extraction due to the fact that, while they are written in different styles, they deal with the same topic. This guarantees a higher lexical variety and a wider coverage of the relation between argument structures and denoted event mentions.\smallskip

\noindent \textbf{Memoirs}. This subcorpus consists of 25 books, historical memoirs of Italian partisans from World War II in North-Western Italy. The time span of the depictions ranges from the 8 September 1943 to 25 April 1945, a period known in the Italian historiography as ``Resistenza'' (Resistance).  Out of 25 books, 20 have been obtained by manual digitization from the original printed editions, while the remaining 5 have been acquired through automatic conversion from existing digital editions. The digitization has been performed first by scanning the original sources and then by automatic conversion to text using an OCR software (Adobe Acrobat Pro DC, version 2015). Despite the good performance of the employed OCR, a subsequent manual cleaning has been necessary. The definition of the corpus, digitization and manual cleaning thereof required about 2 man-months. This acquisition effort resulted in a textual corpus of $\approx$1.5 million words and over 95,000 sentences.\smallskip

\noindent \textbf{Biographies}. The second subcorpus has been obtained from the Wikisource page dedicated to ``Women and Men of the Italian Resistance'',\footnote{ \url{https://it.wikisource.org/wiki/Donne\_e\_Uomini\_della\_Resistenza}} which contains short biographic records of over 3,000 persons involved in the Italian Resistance provided by the National Association of Italian Partisans (ANPI). We kept all entries related to persons who also appear in the Memoirs corpus, for a total of 189 biographies, 57,400 words and 2,500 sentences.\smallskip

\noindent \textbf{Wiki-Articles}. The last subcorpus has been created by collecting 1.748 articles from the Italian Wikipedia, corresponding to the category ``Resistance movements during World War II''.\footnote{ \url{https://it.wikipedia.org/wiki/Categoria:Movimenti\_di\_resistenza\_della\_seconda\_guerra\_mondiale}} This large subcorpus counts $\approx$1.3 million words and 53,000 sentences. The category groups together all the entries connected to resistance movements against nazifascism during World War II, so the Italian Resistance movement (of which the Memoirs subcorpus only covers a minor part) is a subset of it. Additionally, the Wiki subcorpus is far more heterogeneous than the previous one, as the documents span over a range of thematic categories (literature, cinema, historical events).\smallskip

\noindent \textbf{Memoirs-test}. In order to evaluate the system and the event-predicate dictionary (see Section 3.2) on the task of Event Extraction, a test corpus has been created employing memoirs that we have initially excluded from the Memoirs collection. It consists of by 112,000 words and 5,100 sentences.

\begin{table}
\caption{Composition of the textual collection.}
\begin{tabular}{lrrrc}
 & Documents & Words & Sentences & Acquisition\\ \hline
Memoirs & 25 books & 1,469,000 & 95,500 & Digitization\\
Bio & 189 entries & 57,400 & 2,500 & Web scraping\\
Wiki & 1,748 articles & 1,364,000 & 53,800 & Web scraping\\ \hline
Mem-test & 3 books & 112,000 & 5,100 & Digitization \\
\end{tabular}

\end{table}

\subsection{Knowledge Resources}
Beside the textual corpus, the pipeline for Event Extraction requires a set of knowledge resources; each of them is used in our work for a specific task. Below we provide an overview of the employed knowledge resources and clarify their role in the pipeline, which will be described in more detail in Section 5.\smallskip

\noindent \textbf{Gazetteers.} The majority of the books belonging to the Memoirs sub-corpus are accompanied by a list of names of the people mentioned. The partisans are often paired with their related nickname, or \textit{nom de guerre}. We manually merged these lists and, during the digitization of the corpus, we extended them to consider also the specific Locations and Organizations present in the text (with their related abbreviations and/or acronyms). As we have already shown in a previous work \cite{rovera2017domain}, employing a specific knowledge resource for this task is necessary due to the lack of domain coverage of general purpose knowledge bases derived from Wikipedia. The gazetteers are used in the pipeline for extraction and disambiguation of Named Entities and consist of 3,041 Persons, 1,725 Locations and 245 Organizations.\smallskip

\noindent \textbf{Semantic Types Dictionary.} Based on our background knowledge of the domain, a set of representative nominal words is assigned to each of the 25 Semantic Types (see Section 4 for more details). The number of words for each type varies, ranging from 68 for the \textsc{HUMAN\_COLLECTIVE} type to 1 for the \textsc{WATERFLOW} type. This variation also accounts for the lexical variety and richness encoded by each category. For example, the \textsc{HUMAN} semantic type represents not only common words designating a human being like \textit{signora}, \textit{uomo} or \textit{ragazza},\footnote{ ``lady'', ``man'', ``girl''.} but also words referring to social roles like \textit{capitano}, \textit{madre}, \textit{attivista}, \textit{carceriere},\footnote{ ``captain'', ``mother'', ``activist'', ``jailer''.} which denote human beings through the role they have in a social context. This small lexical resource is composed of 424 words and is employed in the semantic tagging step, discussed in Section 5.\smallskip

\noindent \textbf{A dictionary of event-predicate lexical patterns.} The most relevant knowledge resource for the aim of this work is the dictionary of event-predicate lexical patterns, exemplified in Table 2 and Table 3. The main purpose of the dictionary is to enable, in a single step, the recognition and classification of events, the extraction of event participants, and the labeling of each detected participant with the semantic role it plays in the event. This resource is partially inspired by the Corpus Pattern Analysis (CPA) methodology \citep{hanks2004corpus, hanks2005pattern}, which aims at assigning a sense to a verbal or nominal root using a syntactic-semantic pattern of its arguments, derived from corpus evidence. On the other side, the dictionary is also inspired by FrameNet, in that it provides a way to link a set of linguistic terms to a concept (events, in our case) and to retrieve the structure of such concept in texts. As opposed to CPA, \emph{a)} our event-predicate dictionary is not meant to map lexical senses but event mentions denoted by linguistic terms, \emph{b)} each argument pattern is described in isolation and later associated to an event class in which it can appear, and \emph{c)} a syntactic-semantic argument pattern for a given term may belong to (or denote) different event classes. For example, the pattern \textit{abbandonare}\footnote{ ``to leave''.} :: \texttt{subject} :: \textsc{PER} could belong to at least two event classes, namely \textsc{DEPARTING} and \textsc{QUIT\_GROUP}; depending on the denoted event, the same argument pattern induces two different semantic roles (\textsc{Mover} and \textsc{Member}, respectively). Therefore, for each annotated argument pattern, the resource also provides the semantic role typically associated with it for the given event class. The resource has been created through manual analysis of the syntactic-semantic argument structures of a given set of lexical units in the whole corpus (Memoirs, Biographies, Wiki-Articles). The event-predicate dictionary counts 246 verbal or nominal roots, mapped to 88 event classes: 124 verbs (like \textit{aderire}, \textit{imprigionare}, \textit{partire}\footnote{ ``to join'' (an organization), ``to imprison'', ``to depart''.}), 77 nouns (e.g., \textit{arresto}, \textit{liberazione}, \textit{arrivo}\footnote{ ``arrest'', ``liberation/release'', ``arrival''.}), 45 multiword verbal expressions (\textit{aprire il fuoco}, \textit{fare prigioniero}\footnote{ ``to open fire'', ``to take prisoner''.}). The 88 event classes cover three high-level event categories: conflict events, movements of persons and artifacts, and events concerning the membership of individuals to organizations.

\section{Modeling events}
Events, along with their participants, are the conceptual notion that we model in order to recognize their mentions in texts. Given the immense relevance that events have in the domain of historical research \citep{sprugnoli2017one}, in this work we decided to approach three different types, namely spatial events (movements), conflict events and membership in organizations. The setting is inspired by FrameNet (FN) \citep{baker1998berkeley, ruppenhofer2016framenet}, with the notable difference that we are only interested in modeling event information, while many FN Frames also model other type of information.\footnote{ From an ontological point of view, Frames describe ``types of situations'' \citep{ovchinnikova2010data}.}

With this in mind, in our work we denote a textual event mention by the following combination of features:\smallskip

\begin{enumerate}
\item a lexical unit (LU);
\item a set of syntactic dependencies associated to the LU (subject, object, etc.);
\item a set of possible semantic types as fillers of each dependency (e.g. \textsc{HUMAN\_COLLECTIVE}, \textsc{PLACE}, \textsc{VEHICLE}, etc.); 
\item a set of possible semantic roles that can be assigned to each combination of LU - dependency - semantic type (for example \textsc{Mover}, \textsc{Source}, \textsc{Victim}, etc.);
\item an event class that identifies the specific event type.\smallskip
\end{enumerate}

The basic assumption is that the syntactic argument structure of a LU, at least for verbal and multi-word verbal expressions, is the main source of information about the participants of an event denoted by that LU. It follows that there is a correspondence between the fillers of certain syntactic dependencies of a LU and the participants to the event. If we consider the example in Figure 1: the main verb 
\textit{arrestare} (to arrest) is the head of three syntactic dependencies, each of which is the head of a phrase representing a participant in the denoted event. Thus we obtain:
\begin{table}[ht]
\begin{tabular}{lllll}
Duccio Galimberti & \texttt{nsubj} & PER & & \textsc{Arrested}  \\
Torino & \texttt{nmod} & [a [LOC]] & & \textsc{Place} \\
29 novembre 1944 & \texttt{nummod} & DATE & & \textsc{Time} \\
\end{tabular}
\end{table}

\noindent \textbf{1. Lexical Units}. The term lexical unit is used in FrameNet to describe words whose occurrence in text trigger a given frame. Although from a linguistic point of view events can be referenced in text by words belonging to different parts of speech \citep{sauri2009annotating}, in this work we reduce the focus only to verbs, nouns and multi-word verbal expressions. This choice is motivated by the evidence that these categories of words have explicit syntactic relations (dependencies) that provide a first basic structure for the referenced event and participants. As opposed to this, other lexical categories (notably adjectives and adverbs), which often express entity properties, do not offer this syntactic richness in Italian. Like in FrameNet, each LU can map one or more event types.\smallskip

\noindent \textbf{2. Syntactic Dependencies}. For events denoted by verbs and by multi-word verbal expressions we consider the active/passive subject (\texttt{nsubj}, \texttt{nsubjpass} types in Universal Dependencies\footnote{  \url{http://universaldependencies.org/it/dep/}}), the direct object (\texttt{dobj}), all the nominal modifiers (\texttt{nmod}), which map indirect complements like temporal, spatial and others and the numeric modifiers (\texttt{nummod}), usually referring to temporal or quantified complements. For nouns, only \texttt{nmod} dependencies are considered. Although this set of dependency types does not account for all the possible pieces of information related to the event mentioned in a sentence (especially where the sentence is constituted by more than one clause), it represents a stable syntactic structure that conveys most of the relevant information.\smallskip

\noindent \textbf{3. Semantic Types}. Semantic types are used to label the head of lexical arguments and provide information about the entity that appears as filler of a given syntactic dependency.\footnote{ The set of Semantic Types considered in this work is depicted in Figure 3}
While some types are more general (\textsc{HUMAN}, \textsc{HUMAN\_COLLECTIVE}, \textsc{POPULATED\_PLACE}), others are specific to the domain under study (\textsc{POST}, \textsc{WEAPON}). In addition, named entity types \textsc{PER}, \textsc{LOC}, and \textsc{ORG} are included as Semantic Types, as well as the four TIMEX3 temporal tags \textsc{DATE}, \textsc{TIME}, \textsc{DURATION}, and \textsc{SET}.\footnote{ \textsc{PER}, \textsc{LOC} and \textsc{ORG} are standard named entity labels derived by using the TINT pipeline \citep{aprosio2016italy}. TIMEX3 labels are established temporal tags used for instance by systems like HeidelTime \citep{strotgen2013multilingual, strotgen2014extending} and offered by TINT as temporal tagging component.} By means of these three components (LU, syntactic dependency, semantic type) it is possible to build syntactic-semantic patterns associated to a LU (see Tables 2 and 3 for examples).\smallskip

\begin{table}
\caption{Example of syntactic-semantic argument patterns associated to the lexical unit \textit{bombardare} (to bomb).}
\begin{tabular}{l}
\textbf{\textit{bombardare}} (to bomb)\\
\texttt{nsubj} :: \textsc{ORG}\\
\texttt{nsubj} :: \textsc{HUMAN\_COLLECTIVE}\\
\texttt{nsubjpass} :: \textsc{LOC}\\
\texttt{dobj} :: \textsc{INFRASTRUCTURE}\\
\texttt{nmod} :: [da [\textsc{VEHICLE}]]\\
\texttt{nmod} :: [con [\textsc{WEAPON}]]\\
\end{tabular}
\end{table}

\begin{table}
\caption{Excerpts from the mapping of two lexical units to the correspondent event classes. In each event class, all syntactic-semantic argument patterns associated to the verb are associated to a semantic role.}

\begin{tabular}{clll}
Lexical Unit & Event Class & Syn-Sem pattern & Semantic Role \\
\hline
\textbf{\textit{abbandonare}}& &  &\\
(to leave) & & &  \\
& \textsc{DEPARTING} & & \\
 & & \framebox{\texttt{nsubj} :: \textsc{PER}} & \textit{\textsc{Mover}} \\
 & & \texttt{dobj} :: \textsc{LOC} & \textsc{Source} \\
 & & \texttt{nsubj} :: \textsc{ORGANIZAT.} & \textsc{Mover}  \\
 & & \texttt{nsubj} :: \textsc{GROUP} & \textsc{Mover}  \\

 & \textsc{QUIT\_GROUP} & &  \\
 & & \framebox{\texttt{nsubj} :: \textsc{PER}} & \textit{\textsc{Member}}  \\ 
& & \texttt{dobj} :: \textsc{ORGANIZAT.} & \textsc{Group}  \\

\textbf{\textit{abbattere}} & & &\\
(to tear down / to kill) & & &  \\ 
& \textsc{DESTROYING} & &  \\
 & & \texttt{nsubjpass} :: \textsc{INFRASTR. } & \textsc{Patient} \\
 & & \texttt{dobj} :: \textsc{INFRASTR.} & \textsc{Patient}  \\
 & \textsc{DOWNING} & &  \\
 & & \texttt{nsubjpass} :: \textsc{VEHICLE } & \textsc{Patient}  \\
 & & \texttt{dobj} :: \textsc{VEHICLE} & \textsc{Patient} \\ 
& & \texttt{nsubj} :: \textsc{HUMAN} & \textsc{Agent}  \\
 & & \texttt{nsubj} :: \textsc{ORG} & \textsc{Agent} \\
 & \textsc{KILLING} & &  \\
 & & \texttt{nsubj} :: \textsc{PER} & \textsc{Killer}  \\
 & & \texttt{nsubjpass} :: \textsc{PER} & \textsc{Victim}  \\
 & & \texttt{dobj} :: \textsc{PER} & \textsc{Victim} \\
 & & \texttt{nmod} :: [con [\textsc{WEAP.}]] & \textsc{Instrum.} \\
 & & \texttt{nsubj} :: \textsc{ORG} & \textsc{Killer}  \\
\end{tabular}
\end{table}

\noindent \textbf{4-5. Semantic Roles and Event Classes}. We employ as event classes a set of FrameNet frames and as semantic roles the corresponding Frame Element labels. We extend both of them in order to fully describe the collection under study and account for relevant event types for the domain at hand. We use overall 88 event classes (as described in 3.2) of which 52 correspond to existing FrameNet frames, while 36 are specific for modeling event types present in our collection. The latter concern in particular {a)} events that are not modeled by any FrameNet frame, for instance specific war-related event classes like \textsc{DEPORTATION}, \textsc{LIBERATION} or \textsc{AIRDROP} as well as fine-grained movements, like \textsc{GET\_OFF\_VEHICLE} or \textsc{ENTER\_BUILDING}; {b)} events that are modeled by some FrameNet frame but in a very general way, which would not account for its relevance in our corpus, for instance \textsc{DISARM}, modeled in FN by the \textsc{EMPTYING} frame, the \textsc{BOMBING} event, modeled in FN as an \textsc{ATTACK} frame, or still the \textsc{RETREAT} event, modeled generally as \textsc{QUITTING\_A\_PLACE}. From the perspective of the overall goal of this work, i.e. providing fine-grained access to a historical collection of war memoirs, this is the most important step as it allows us to assess which type of event is mentioned in a given context, to associate a set of entities (Named Entities or other types of entities) to that event, and to assess which role is played, with respect to the event, by each event participant (example in Table 3).

\section{Description of the system}

In this section the implementation of our system for Event and Participants Extraction is described. The pipeline (sketched in Figure 2) is composed of three macro-steps: \emph{1)} extraction and disambiguation of Named Entities, \emph{2)} tagging of lexical units' arguments with semantic types, \emph{3)} recognition of events and their participants. Tasks \emph{1)} and \emph{2)} are described in Section 5.1, while task \emph{3)} in Section 5.2. As general-purpose NLP pipeline we use TINT \cite{aprosio2016italy}, an open source NLP pipeline for Italian based on Stanford CoreNLP. This software is employed for all pre-processing tasks (tokenization, POS-tagging, dependency parsing and NER).

\begin{figure}
\includegraphics[width=1.0\linewidth]{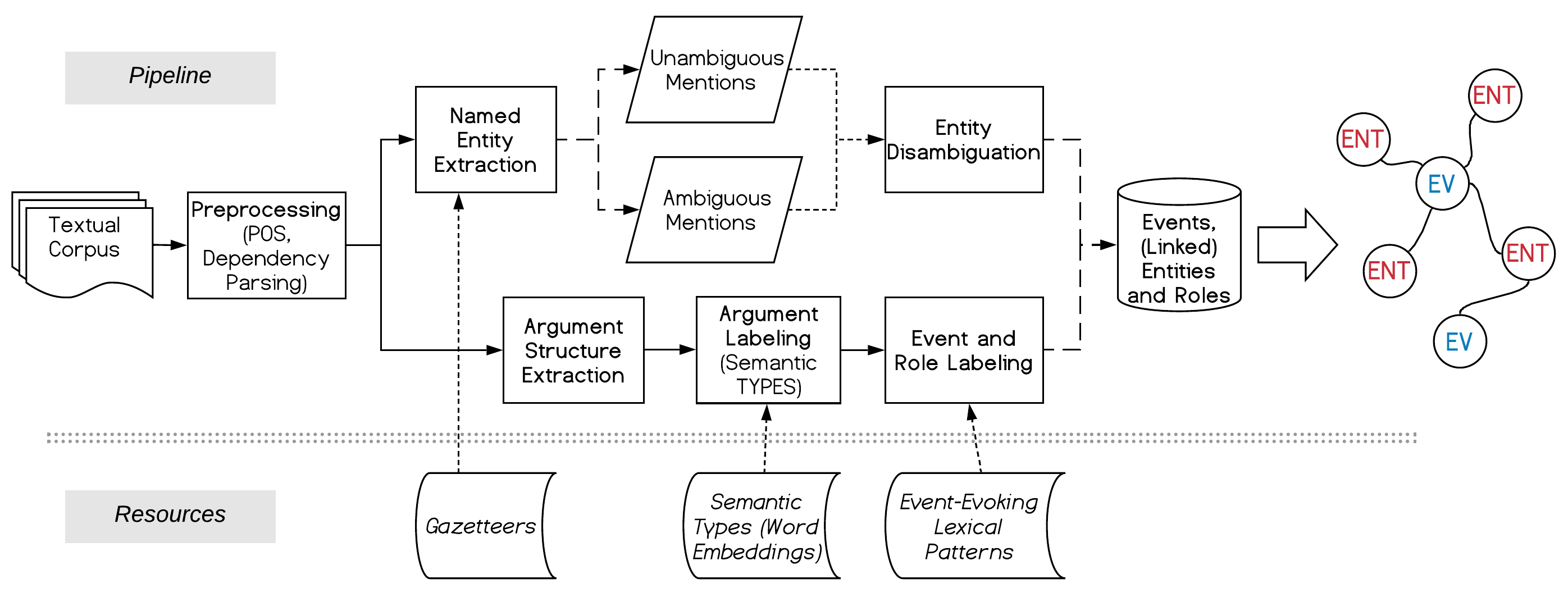}
\label{centroid}
\captionof{figure}{The full text-to-knowledge pipeline described in this paper.}
\end{figure}

\subsection{Named Entity Extraction and Disambiguation} 

The first step of our pipeline consists of extracting all mentions of Named Entities (Persons, Locations and Organizations) and linking them to the respective unambiguous entries in the gazetteer; to do so, we employ the following strategy.

\subsubsection{Surface form recognition} For Locations and Organizations we identify, for each entry and for each lexical variation listed in the gazetteer, the corresponding textual mentions. For entities of type Person the look-up process is slightly more complex since persons in our collection are often referred to by different combinations of their name, surname and nickname (if any). Therefore, we first pre-process the Person gazetteer and produce, for each entry, a list of lexical patterns by combining these three building elements;\footnote{ For example: ``Name Surname Nickname'', ``Name Nickname Surname'', ``Nickname Surname'', etc. We identified 26 frequent patterns.} then, for each Person entry, we identify in text all potential mentions.\smallskip

\subsubsection{Disambiguation and Linking} 
All the identified textual mentions that correspond to a unique entry in the respective domain gazetteer, i.e., all non-ambiguous mentions, are directly linked to the knowledge base. For all the remaining mentions, that is, for all those textual mentions that correspond to more than one entry in the gazetteers, a disambiguation step is required for them to be linked to the correct entry. While for entities of type Location and Organization ambiguity in our corpus is more or less negligible, for Persons it is highly relevant. In particular, the following two types of ambiguity need to be resolved: cross-classes and intra-classes.\smallskip

\noindent \textbf{Cross-classes ambiguity}. In a certain number of cases, a given surface form can refer to entities, which may belong to more than one entity class. This is often the case for persons’ surnames that are also
toponyms (\textit{Genova}, \textit{Alessandria}, \textit{Siracusa} are some examples) or for organizations which took their name from fallen fighters (\textit{Brigata Rolando Besana}) or activists (\textit{Brigata Carlo Rosselli}). Although the phenomenon is not pervasive - it affects around 1000 instances in the Memoirs corpus -, it is a feature of our textual corpus and it causes a systematic drop of performance. Moreover, the issue often arises when a single-token surface form is used, like in the sentence \textit{Ai reparti della Besana toccò schierarsi sulla destra del dispositivo}\footnote{``The units of the ``Besana'' had to be deployed to the right.''}.
In order to tackle this problem, we used the sentences annotated by the previous direct linking phase - given their high precision - to train a classifier that is able to assign the correct label (\textsc{PER}, \textsc{LOC}, \textsc{ORG}) to the multiclass ambiguous mentions. Four different classifiers have been tested: K-Nearest Neighbour, Nearest Centroid, Naive Bayes, Support Vector Machine. The set of features used for training is the following:
\begin{itemize}
\item[] 1. the word preceding the Named Entity (NE) occurrence;
\item[] 2-3. the part of speech of the 2 words preceding the NE occurrence;
\item[] 4-5. the part of speech of the 2 words following the NE occurrence;
\item[] 6. the average word embedding\footnote{ We expand on the use of word embeddings in our work in the next sub-section.} of the 2 words before the NE occurrence;
\item[] 7. the type of dependency linking the NE to the verb;
\item[] 8. the type of verb linked to the NE (movement, social, conflict).\smallskip
\end{itemize}
The learning algorithm achieving the best results in prediction was the k-NN (k = 9), scoring 0.75 F1-score (macro) on a sample of 250 manually annotated sentences. We also observed that features 2, 3 and 7 are most relevant for the task.\smallskip

\noindent \textbf{Intra-class ambiguity}. In order to deal with intra class ambiguity (which affects the most number of cases), we rely on the basic assumption from the entity linking literature \cite{witten2008effective,shen2015entity} that co-occurrence is a valuable source of information for determining the identity of an entity mention. We build an iterative process where each ambiguous mention  is resolved by considering how frequently each of the possible candidates appears with the other (already disambiguated) surrounding mentioned entities in the same sentence. First, we compute an association score between each pair of disambiguated entities in the whole corpus (Memoirs) in the following way:
\begin{displaymath}
association =\frac{co\_occur(\textit{A},\textit{B})}{(\frac{f\textsubscript{A} + f\textsubscript{B}}{2})}\smallskip
\end{displaymath}
\noindent where \textit{co\_occur}(\textit{A}, \textit{B}) represents the co-occurrence, in absolute terms, between the two entities A and B, while the denominator represents the averaged sum of their overall frequency \footnote{We also experimented using Pointwise Mutual Information for this task, achieving slightly worse results for the whole task.} \footnote{This measure is characterized by several features: it \textit{a)} is symmetric, \textit{b)} ranges from 0 (no association) to 1 (perfect association, the two entities always appear together in the corpus), \textit{c)} uses the average at denominator to “normalize” the score
according to the magnitude of the two frequencies. This is useful when the frequencies have different size orders and \textit{d)} it is biased towards low frequency entities, in that it tends to overestimate association when both entities have a low absolute frequency; this is a desirable feature in our use case as we have a long tail of sparse entities, but if needed it can be circumvented by applying a cutoff threshold.}. Then, given an ambiguous mention, we rank each candidate for that mention based on the association score between the candidate and all the disambiguated entities appearing in context. The ambiguous mention is then linked to the candidate achieving the highest score.
Let us consider the following example given the verb \textit{rientrati} (they came back):\smallskip

\begin{quote} \textit{19 marzo 1945 [ ... ] Sono \underline{rientrati} Renato, Saro, Nino, Pino, Marco, Carlin e Siracusa con 26 (sic) uomini, non cattivi, ma non dei migliori [...]}.\footnote{ ``19th March 1945 [...] Renato, Saro, Nino, Pino, Marco, Carlin and Siracusa came back with 26 (sic) men, not bad ones, but not the best [...]''.\smallskip}
\end{quote}

As a first step, the cross-class ambiguity of the argument \textit{Siracusa} (both a city name and the nickname of a partisan) will be resolved using the previously discussed method (k-NN). Having recognized it as a named person present in our gazetteer, we will use this information to disambiguate the other 6 names considering the following numbers of candidates: \textit{Renato} (15 candidates), \textit{Saro} (2 candidates), \textit{Nino} (25 candidates), \textit{Pino} (8 candidates), \textit{Marco} (14 candidates), \textit{Carlin} (4 candidates). In the evaluation section we provide evidence that addressing intra class ambiguity this way permits us to identify over 3,000 additional disambiguated entity mentions.

For all the remaining named entities not listed in our gazetteers as well as for time expressions, we integrate in our system the output of the NER module of the TINT pipeline, which tags them as \textsc{PER}, \textsc{LOC} and \textsc{ORG}.  

\subsection{Semantic Type Classification} NEs represent only a subset of the semantic types of lexical fillers in the argument structure of a LU. To identify the type of the other arguments related to a given syntactic head, we combine our dictionary of semantic types (see 3.2) with word embeddings \cite{mikolov2013distributed}, a relatively recent computational linguistic technology grounded on the distributional hypothesis \citep{harris1954distributional}. We use 300 dimensional pre-trained fastText embeddings\footnote{ \url{https://dl.fbaipublicfiles.com/fasttext/vectors-crawl/cc.it.300.bin.gz}} \citep{bojanowski2017enriching} in the following way:  

\begin{enumerate}

\item We start by creating a centroid for each semantic type in the dictionary (\textsc{BUILDING}, \textsc{VEHICLE}, \textsc{HUMAN\_COLLECTIVE}, etc.) as the averaged sum of the word embedding vectors of the nominal words belonging to that type; this centroid represents the ``center of mass'' of each semantic type. A visual representation of the obtained centroids is provided in Fig. 3.\footnote{ The plot has been produced using t-SNE \citep{maaten2008visualizing}, perplexity=2, 5000 iterations.}\smallskip

\item In the same way, we represent each new argument to be tagged through the word embedding vector corresponding to its head. \smallskip

\item Finally, arguments are tagged by computing the cosine similarity between their semantic vector and each of the centroids and assigning it to the closest one, as in a Rocchio classifier \cite{schutze2008introduction}.
\end{enumerate}
\begin{figure}
\includegraphics[trim={4cm 4cm 3cm 0}, width=0.75\linewidth]{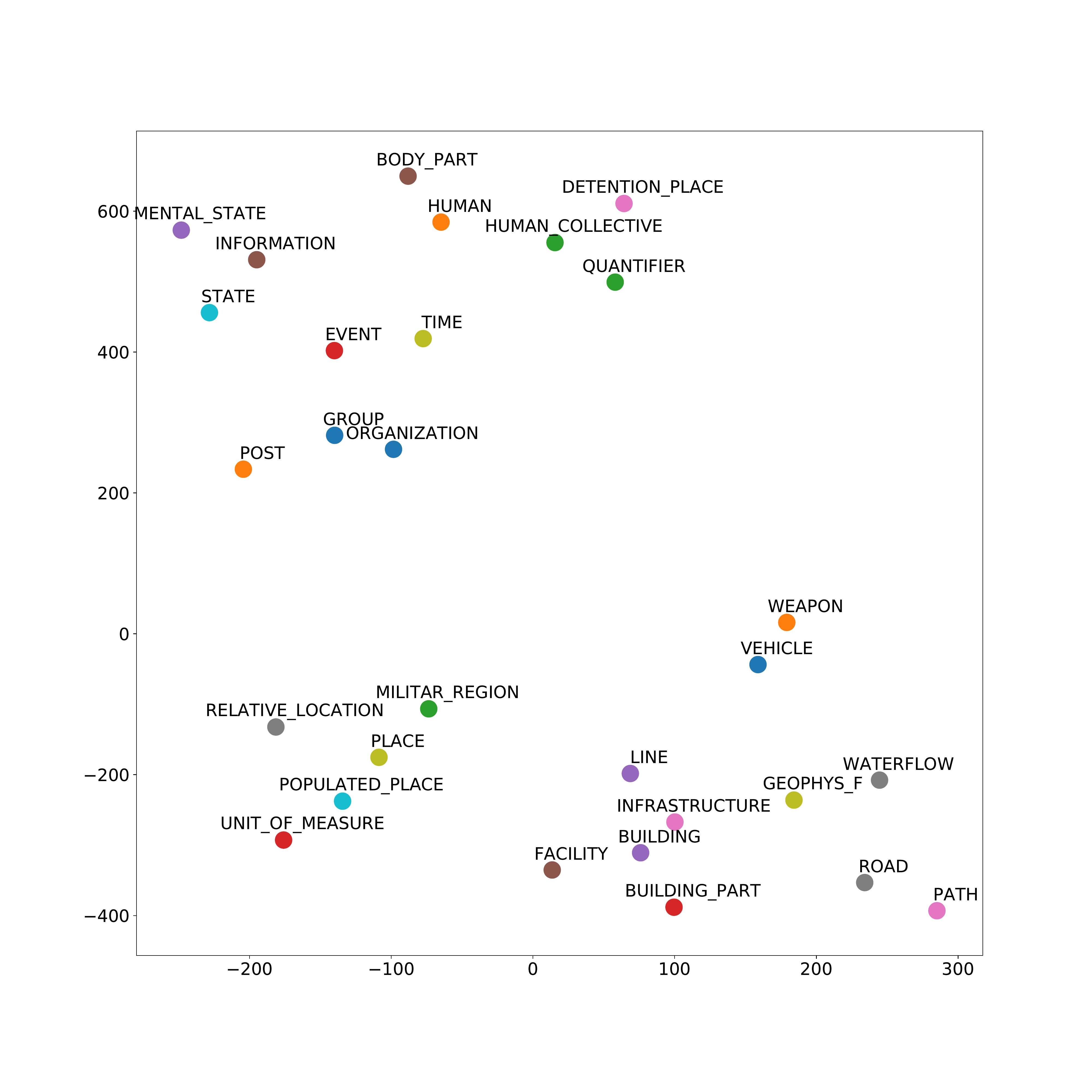}
\label{centroid}
\captionof{figure}{A plot in two dimensions of the centroids representing semantic types (created with t-SNE).}
\end{figure}

\subsection{Event classification and Role Labeling}
Once arguments are labeled with semantic types, we have all elements in place to use our event-predicate dictionary. Let us consider the following example:\smallskip

\begin{quote}
\textit{Sempre negli stessi giorni dell'11 e 12 settembre, da Pinerolo \underline{salirono} a Barge (in Valle Po) alcuni ufficiali di Cavalleria.}\footnote{ ``In the same days, 11th and 12th September, some cavalry officiers came up to Barge (in the Po Valley) from Pinerolo.''\smallskip}
\end{quote}

The final goal of our pipeline is to classify the event triggered by the lexical unit \textit{salirono} (they came up) according to the available event types and to assign a semantic role to each of the tagged arguments of the event anchor. To do so, we follow this procedure:

\begin{enumerate}
\item The sentence is represented as a set, consisting of the lexical unit and the annotated arguments, each argument labeled with its syntactic dependency, semantic type and preposition, as in Table 4.

\begin{table}[H]
\caption{lexical unit, arguments, syntactic dependencies and semantic types for the sentence in the example.}
\begin{tabular}{ll|l}
Lexical unit: & \textit{salire} & \\
Tagged args: & \textit{ufficiali} & \texttt{nsubj} :: [\textsc{HUMAN\_COLLECTIVE}] \\
 & \textit{Pinerolo} & \texttt{nmod} :: [a [\textsc{LOC}]] \\
 & \textit{Barge} & \texttt{nmod} :: [da [\textsc{LOC}]] \\
 & \textit{Valle Po} & \texttt{nmod} :: [in [\textsc{LOC}]]\smallskip \\
\end{tabular}
\end{table}

\item Using our resource, all the possible event classes corresponding to the given lexical unit are retrieved. In this example, the LU \textit{salire} can trigger three classes: \textsc{BOARD\_VEHICLE}, \textsc{MOVE\_UPWARDS} and \textsc{PATH\_SHAPE};\smallskip

\item Each event class is described in the event dictionary by a set of lexico-syntactic argument patterns; the set of tagged arguments from the sentence is compared with the sets of each candidate class and the class scoring the highest intersection with the tagged argument structure of the sentence is assigned to the LU. In the example, the result is the following:\smallskip

\begin{description}
\item \textsc{MOVE\_UPWARDS}: 4
\item \textsc{PATH\_SHAPE}: 2
\item \textsc{BOARD\_VEHICLE}: 1
\end{description}

The sentence is thus assigned to the \textsc{MOVE\_UPWARDS} type.\smallskip

\item Finally, the semantic roles provided by the event class are assigned to each argument, which delivers a full-round semantic representation of the event mention, as presented in Table 5.\smallskip

\begin{table}
\caption{Sentence annotated with event type and semantic roles.}
\begin{tabular}{ll|ll}
Lexical unit: & \textit{salire} & & \\
Tagged args: & \textit{ufficiali} & \texttt{nsubj} :: [\textsc{H.\_COLLECT.}] & \textbf{\textsc{Mover}}\\
 & \textit{Pinerolo} & \texttt{nmod} :: [a [\textsc{LOC}]] & \textbf{\textsc{Source}}\\
 & \textit{Barge} & \texttt{nmod} :: [da [\textsc{LOC}]] & \textbf{\textsc{Goal}}\\
 & \textit{Valle Po} & \texttt{nmod} :: [in [\textsc{LOC}]] & \textbf{\textsc{Goal}}\\
\hline
\textbf{Event Class} & & \textbf{\textsc{MOVE\_UPWARDS}} & \\
\end{tabular}
\end{table}
\end{enumerate}

Two exceptions to our standard pipeline need to be discussed further: \emph{a)} often the given lexical unit maps only to one event class. In this case, we classify the event and assign roles if at least one tagged argument from the argument structure of the sentence matches the set of arguments provided by the event class (in other words, if in the above mentioned procedure, it scores at least 1 in step 3). \emph{b)} In case of ties between two or more classes, we currently do not tag the event as for our final application we value precision over recall\footnote{The reason for this is twofold. First, in a rule-based system like ours, Recall performance is mostly due to the richness of the lexical pattern dictionary, which is by definition always extensible. At this stage, we aim to understand if and to which extent the patterns are productive. Second, since we envisage to use the data for visual representation/querying, we prefer to provide precise result, at the cost of incompleteness.}.
\begin{table}
\caption{Results of event extraction, divided by subcorpus, confidence and type of LU triggering the event.}
\begin{tabular}{lcrrr}
Lexical type & Confidence & \textit{Memoirs} & \textit{Wiki} & \textit{Biographies} \\
\hline
Verbals & High & 3583 & 3151 & 270 \\
& Low & 11823 & 8190 & 454 \\
Nominals & High & 129 & 140 & 8 \\
& Low & 1985 & 3106 & 133 \\
Multi-word verbal & High & 109 & 202 & 12 \\
& Low & 503 & 679 & 55 \\
\hline
& High & 3821 & 3493 & 290 \\
& Low & 14311 & 11975 & 642 \\
\hline
& & 18132 & 15468 & 932 \\
\end{tabular}
\end{table}

Since the system assigns a class label based on the type and number of the tagged arguments of the lexical unit, we suppose that the more arguments of a LU the system is able to tag correctly, the higher the probability is for the class label to be correct. This hypothesis will be substantially confirmed while evaluating the system (see Section 6.3). Therefore, in Table 6 we present the results by keeping high and low confidence results separated. Low confidence mentions are event mentions that have been extracted and classified by the system based on \textit{one single} matching tagged argument, while high confidence mentions are based on at least two tagged arguments.

\section{Evaluation}

In this section we present a quantitative evaluation of each step of our pipeline.
\subsection{Named Entity Extraction and Disambiguation}
The strategy employed for extraction and disambiguation of Named Entities from our corpus consists basically of two steps: first, all non-ambiguous mentions are retrieved and linked through dictionary look-up (which we use as a baseline); then the remaining ambiguous mentions are disambiguated using the method described in Section 5.1.2. In this work we deal with the problem, well known in the Digital Humanities and Digital Library communities, of disambiguating domain-specific entities, for which no entry in general purpose knowledge bases (e.g., DBpedia) is available \citep{munnelly2018investigating,mcdonough2019named,rovera2017domain}. We report the evaluation of the disambiguation strategy based on co-occurrence of entities in text and discuss its limitations. Table 7 shows the results of the evaluation of the system based on a gold standard of 400 sentences manually annotated with Named Entities and their link to the respective gazetteers.

\begin{table}[H]
\caption{Evaluation of Entity Disambiguation on a gold standard of 400 manually annotated sentences (precision-oriented scenario). In brackets, the increase/decrease of performance with respect to the baseline (direct linking of non ambiguous entity mentions).}
\begin{center}
\begin{tabular}{lll}
\multicolumn{3}{c}{PER} \\ \hline
Precision & 0.842 & (-0.056) \\
Recall & 0.669 & (+0.115) \\
F1 score & 0.746 & (+0.078) \\
Linked mentions & 18709 & (+3306) \\
Ambiguous mentions & 11791 & \\
 & & \\
\multicolumn{3}{c}{LOC} \\ \hline
Precision & 0.936 & (+0.000) \\
Recall & 0.927 & (+0.007) \\
F1 score & 0.931 & (+0.003) \\
Linked mentions & 26761 & (+257) \\
Ambiguous mentions & 959 & \\
 & & \\
\multicolumn{3}{c}{ORG} \\ \hline
Precision & 0.931 & (+0.000) \\
Recall & 0.911 & (+0.000) \\
F1 score & 0.921 & (+0.000) \\
Linked mentions & 4943 & (+34) \\
Ambiguous mentions & 34 & \\
\end{tabular}
\end{center}
\end{table}
For each entity type, we report Precision and Recall of the disambiguation/linking task, along with the absolute number of linked mentions and the remaining ambiguous mentions. For Locations and Organizations the performance is fairly high and the disambiguation step does not improve significantly the score on the baseline. This is due to the low intrinsic ambiguity of these entity types. Concerning Persons, in the chosen setting, our disambiguation strategy allows to improve the Recall of more than 11 points by losing around 5 points in Precision, leading to an overall increase of the F1-score of almost 8 points.\smallskip

\noindent \textbf{Error Analysis} (Persons). Through the analysis of false positives and false negatives we can dig deeper into the causes of such errors. Regarding Precision, it turns out that only 48\% of errors are due to a wrong classification of the system, while the remaining 52\% are represented by cases of homonymy either caused by entities not present in the gazetteers or by mismatches between Named Entities and other parts-of-speech. Where Recall is concerned, we observed that \emph{a)} more than 90\% of false negatives occur with single-token mentions, mainly first names or surnames and, more importantly, that \emph{b)} the mentions are isolated, i.e., the context in terms of other disambiguated mentions is very poor. This situation prevents the ranking system to work, as all candidates automatically get a zero score. It is also important to note that most of the false negatives occur in texts adopting a diaristic style. As a whole, these observations show that the disambiguation strategy is effective as long as enough contextual information is provided.

\subsection{Semantic Type Classification}
For evaluating the performance of the semantic tagging system described in Section 5.2, a test set of 252 words has been created by randomly picking argument fillers of verbs, nouns and multi-word verbal expressions and by manually annotating them. 
For the evaluation, the \textsc{TIME} semantic type (i.e. centroid) has been added, as in the normal tagging procedure TINT it is used for recognizing temporal expressions. Moreover, an \textsc{OTHER} type has been used in the manual annotated gold standard for labeling words that do not belong to any of the available semantic types. On the automatic side, a threshold of 0.4 in the similarity score has been set, below which a word is tagged as \textsc{OTHER}. Results are shown in Table 8\footnote{Given the small size of the test set, we do not perform a tag-wise evaluation; instead, we report the overall accuracy (i.e., percentage of correctly labeled words).}.\smallskip
\begin{table}
\caption{Evaluation of the embedding-based tagging system on a gold standard of 252 sentences.}
\begin{center}
\begin{tabular}{cccc}
 & Correct & Wrong & Accuracy (\%) \\
\hline
@1 & 190 & 62 & 0.754 \\
@3 & 207 & 45 & 0.821 \\
\end{tabular}
\end{center}

\end{table}
\newline
\textbf{Error analysis.} As shown in the table, the performance of the system is fairly good, but it does not improve significantly when the cut-off rank changes from @1 to @3. The reason for that can be further explained by looking at the incorrectly labeled examples from the test set. By focusing on the evaluation @3, it turns out that in 77\% of the incorrectly labeled cases an \textsc{OTHER} annotation appears, either in the gold standard or in the automatic annotation; this fact suggests that the main source of error is given by the incompleteness of the adopted semantic type system, which only accounts for a limited set of entity types, the ones more relevant for the domain. For example, the adopted semantic types do not account for abstract entities (knowledge, feelings, intentions, etc). The remaining 23\% of incorrectly labeled words (10 cases) represents the genuine source of error, i.e. inter-class error.

\subsection{Event Classification and Role Labeling}
We approach the task of \emph{a)} extracting events and participants and  \emph{b)} labeling participants with roles in different sub-steps. The evaluation of the two tasks is therefore performed separately and it is conducted on the Memoirs-test subcorpus, which was not employed for the creation of the event-predicate dictionary. This corpus has been processed with our pipeline, as described in Section 5; Table 9 summarizes the  number of extracted and classified events.\smallskip

\begin{table}
\caption{Absolute number of anchors and events extracted from the Memoirs-test set.}
\begin{tabular}{lrrr|r}
LU type & Anchors & High & Low & Sum  \\
\hline
verbs & 2002 & 145 & 574 & 719 (73.2\%) \\
nouns & 1224 & 12 & 207 & 219 (22.3\%) \\
multi-word verbs & 87 & 6 & 38 & 44 (4.5\%) \\
\hline
& 3313 & 163 (16.6\%) & 819 (83.4\%) & 982 (100\%) \\
\end{tabular}
\end{table}

\noindent \textbf{Extraction}. For evaluating the extraction performance, we manually annotated 300 sentences randomly chosen from the test subcorpus, among the sentences that contain at least one lexical unit modeled in the event-predicate dictionary. The annotation is binary and assesses whether or not the sentence denotes an event mention. Results are summarized in Table 10. By analyzing the types of error, it turns out that the loss in Precision is mainly due to metaphoric use of language or non-compositionality of the predicate\footnote{Some examples: \emph{respingere una proposta}, ``reject a proposal'', tagged as REPEL; \emph{incontrare resistenza}, ``meet (with) resistance'', tagged as ENCOUNTER; \emph{tornare a perdere il sentiero}, ``lose the trail again'', tagged as RETURN.} (73\% of cases) or to previous errors in the pipeline (27\%). The low Recall has three main reasons: the lack of coverage of the resource, along with the misclassification with different lexical meanings (47\% of cases), the presence of errors in previous tasks of the pipeline, especially in dependency parsing (17\%) and the mention of events that have no relevant syntactic dependencies (17\%), which prevents the system from assigning any class label to the word and to recognise the word as an event trigger at all (see step 3 in Section 5.3).\smallskip
\begin{table}
\caption{Evaluation of the event extraction task on a gold standard of 300 sentences.}
\begin{tabular}{lc}
Precision & 0.78  \\
Recall & 0.50 \\
True negatives & 0.88 \\
F1-score & 0.61 \textbf{}\\
\end{tabular}

\end{table}

\noindent \textbf{Classification}. Given an event mention extracted by the system, we are interested to know how precise our pipeline is at assigning a class label to the mention, i.e. at classifying the event type. For evaluating this step, we randomly collected 200 event mentions extracted by the system, and annotated them manually with the correct event class. The gold standard is further divided in two parts: 100 mentions are chosen among the ``low confidence'' mentions, while the remaining 100 are ``high confidence'' mentions. The results confirm the hypothesis formulated in Section 6.3, since the low confidence set leads to a 0.73 of Precision, while in the high confidence set the precision grows up to 0.89. Overall, these numbers also demonstrate that, when the system is provided with enough correctly extracted semantic information, it performs very well. It should also be considered that this last step is built on top of a long pipeline, thus suffering from a downstream propagation of errors in previous steps.

\section{Representation}
The system described so far produced a knowledge base containing occurrences of events, entities (participants) and qualified relationships (roles) between entities and event mentions such entities participate in.
In order to explore the results of the system and visualize the relationships between entity and event
occurrences, we represent the extracted knowledge as a network of events, participants and roles\footnote{ A Gephi version of the event graph is available for download at \url{https://bit.ly/2UIZISo}}; the resulting graph has two applications:
\begin{enumerate}
    \item providing a structure for querying the event knowledge base;
    \item providing a basic visualization tool.
\end{enumerate}
The network has two types of nodes: event nodes and entity nodes. Since Named Entities (as well as some time expressions) are linked, nodes representing the linked entities appear only once in the graph, while non-linkable entity types (\textsc{HUMAN COLLECTIVE}s, \textsc{VEHICLE}s, \textsc{GEOPHYSICAL FEATURE}s, etc.) appear multiple times, once per mention. Entity nodes are also labeled with the type of the entity and with the corresponding text span. Event nodes are linked to entity nodes by an edge labeled with \emph{a)} the semantic role played by the entity in the event and \emph{b)} the document the event-entity pair has been extracted from.

This structure allows, given a named entity (or, better, given its node in the graph), to directly access all the events in which that entity participates. Moreover, given an event, it is always possible to go back to the document and the sentence where that event is mentioned. 

If we consider the example in Figure 4, one of the events in which the central entity, \emph{Tancredi ``Duccio'' Galimberti} has been involved, was his arrest (``cattura'' - in the bottom left part of the graph). This event is modelled using information extracted from specific sentences in different books, to which we can always go back directly from the graph. For instance we would see how he was captured by the fascists (in the first sentence) and how this event took place in Turin (second):

\begin{quote} \textit{Individuato un recapito delle staffette delle formazioni \underline{i fascisti} vi avevano catturato Galimberti, mentre ritirava le lettere.\footnote{``Having found an address for the formations' couriers, the fascists had captured Galimberti there, while he was collecting the letters.''}}
\end{quote}

\begin{quote} \textit{Un grosso colpo per noi in quel periodo fu la cattura di Duccio Galimberti a \underline{Torino}, catturato davanti alla panetteria dove aveva il recapito clandestino. \footnote{``A big shock for us in that period was the capture of Duccio Galimberti in Turin, caught in front of the bakery where he had his clandestine address.''}}
\end{quote}

As highlighted below, this type of representation offers intermediate access to texts between Close and Distant Reading, by keeping each piece of knowledge (event or entity mention) within its more general context (the whole textual corpus, represented by the whole graph).
By using this representation structure it is possible to benefit from bidirectional semantic access to texts, both from the text to the knowledge (events, entities and roles) and the opposite way from any piece of knowledge to its textual source, with the only constraint being the design choices given by our event model (event classes and semantic roles).

\begin{figure}[H]
\includegraphics[width=0.7\linewidth]{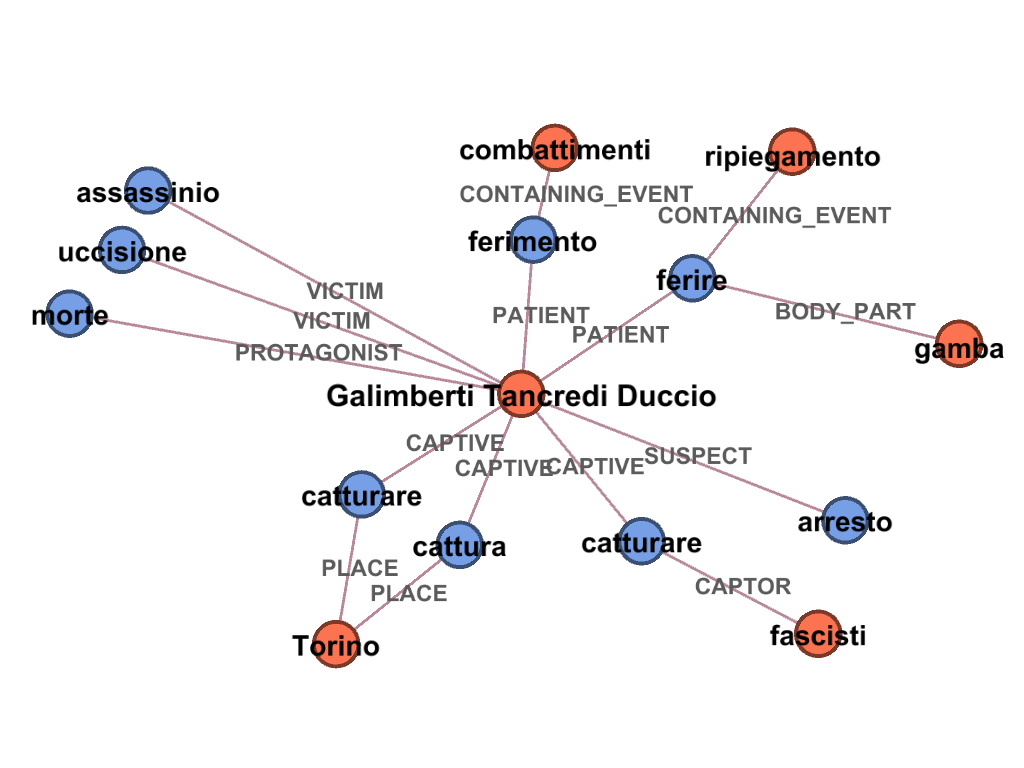}
\captionof{figure}{Injuring, arrest and assassination of ``Duccio'' Galimberti, extrapolated from the event graph. Nodes in blue represent events, nodes in orange represent entities. Edges are labeled with semantic roles.}
\end{figure}

\subsection{Access Scenarios}

\noindent What follows is a brief sketch of three possible access scenarios enabled by our representation structure.\smallskip

\begin{figure}[h!]
\includegraphics[width=0.75\linewidth]{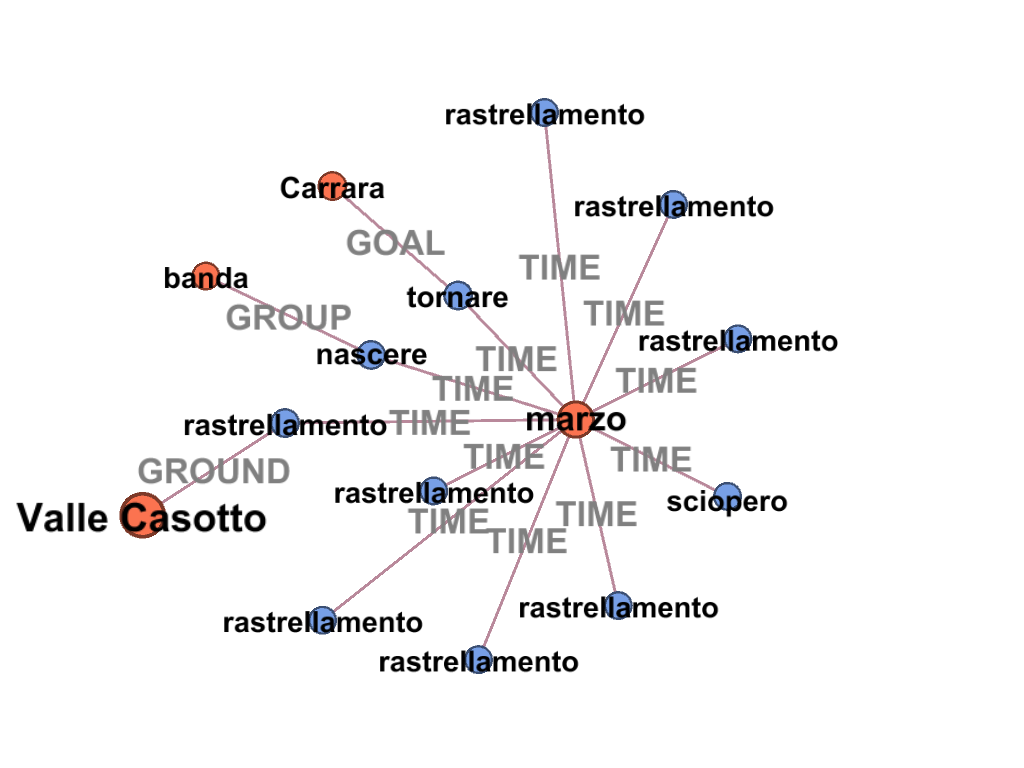}
\captionof{figure}{Time-based egocentric network, depicting events happened in March 1944 (mentions thereof in the Memoirs corpus).}
\end{figure}

\noindent \textbf{Egocentric networks.} An egocentric network (or ego network) is a graph centered on a specific node and it is used to highlight the relationships of the ego node\footnote{An egocentric or ego node is the focal/central node in an egocentric network.} with other nodes. In our case, ego networks can be employed for entity-driven access, by obtaining all events a given entity is involved in. If we return to Figure 4, this depicts all extracted events concerning the person of \textit{Tancredi “Duccio” Galimberti}, a political leader of the Resistance movement. In the top part of the network we can see the event of his injuring (mentioned twice in two different documents), at the bottom is his capture (arrest) in Torino at the hands of fascist militias - mentioned 4 times in 4 documents - and left his murder (3 mentions in 2 documents). Another way to use ego networks is to extract the subgraph related to a given date; for example, Figure 5 shows events happened in the month of March 1944. This access mode is particularly suited for users who already have a focus of interest and need to contextualize it within the dataset.\smallskip

\noindent \textbf{Constrained queries.} Our representation structure can be further leveraged for building advanced queries according to different information needs. By introducing constraints in terms of semantic types and roles we retrieve all entities that fulfill certain roles. For example, people undergoing an arrest can be retrieved by constraining entities on type ``PER'' and ``HUMAN'', events on type ``ARREST'' and semantic roles on ``SUSPECT''. Figure 6 shows a subset of arrest and capture events from the graph. Along this line we can retrieve, for example, deported people, cities or places occupied or liberated or where clashes / battles took place, persons who created or joined organizations, and so forth. This scenario may be useful for users who are interested in a specific event type and want to examine its coverage in the dataset.\smallskip

\begin{figure}[H]
\includegraphics[width=0.75\linewidth]{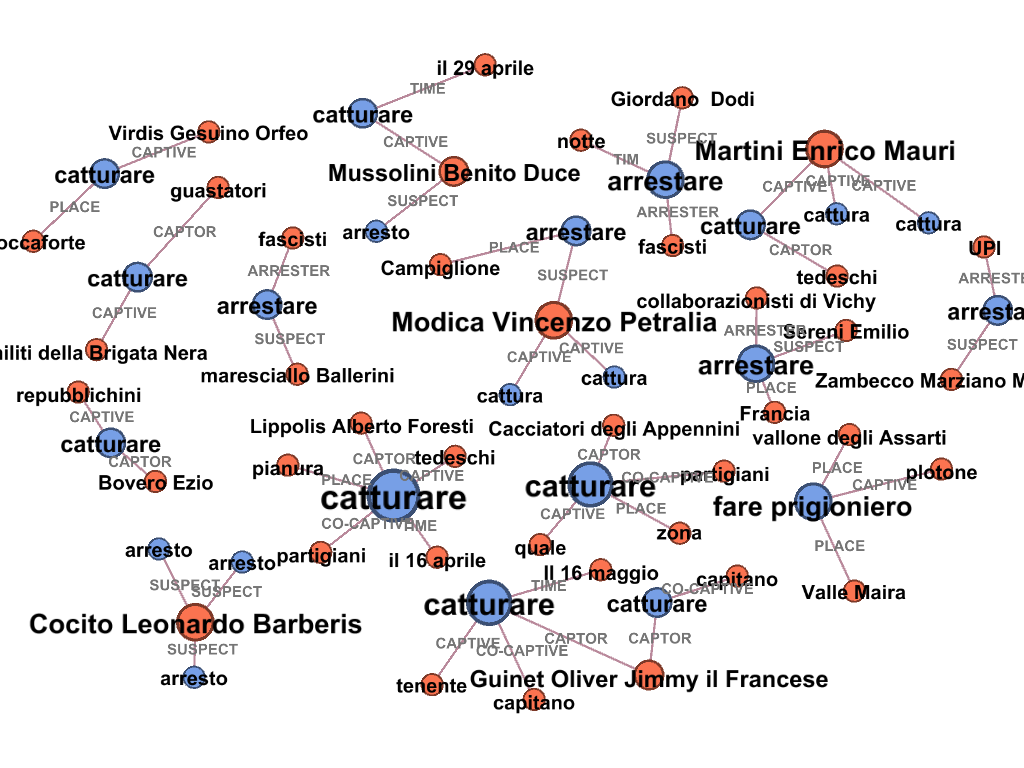}
\captionof{figure}{Arrests and capture events from the graph (sample).}
\end{figure}

\noindent \textbf{Events and entities as summarization tool.} A third scenario supported by our system is the extraction of the subgraph of events and entities related to a specific document or a set of documents. This feature can be seen as a high-level summarization tool and provides the user with a way to compare quickly two documents.

\section{Reusability and Future Steps}

While our work has focused on a highly specific and challenging research domain, the event-extraction pipeline  described above is \textit{per se} domain agnostic and can be adapted to applications in many other scenarios; the domain-dependent component is represented by the event-predicate dictionary (and by semantic types), which should be updated or integrated according to the need. To do this, it could be sufficient to compare the set of our annotated lexical units with the target set of LU of interest in the new domain corpus, in order to assess which LUs and event classes should be updated. The effort required in this step is difficult to predict as it depends heavily on the new domain and on the extraction target; nevertheless, the existing literature on pattern harvesting has shown that in many cases a few patterns are sufficient to cover most occurrences of a given LU (thus drastically reducing the so-called ``lexical entropy'') \cite{hanks2005pattern}. Moreover, by relying on our dictionary of event-predicate patterns, researchers could extract events, participants and roles from Italian news articles, by combining our resource with the output of general-purpose entity linking systems (e.g., TagMe \cite{ferragina2010tagme}), which we could not employ in our work due to the lack of domain coverage. Similarly, our gazetteer could be employed by researchers studying the same historical events, but employing primary sources written by German or American soldiers, due to the fact that the relations between people, places and organizations will stay the same. Finally, this work could be seen as a set of guidelines for researchers aiming to build event-centered networks from historical sources, to whom we highly recommend to start by studying the coverage of resources such as FrameNet for their domain/language and, when lacking, the possibility of modeling the event types under study by extending our relational schema.\smallskip

Based on the experience gained during this work and the results obtained through the network-based visualization, we now envisage three main research directions in the next future:
\begin{enumerate}

\item \textit{Event-predicate lexical dictionary}. The event-predicate dictionary has proved to be a very useful and effective knowledge resource for extracting information from text. Since it partially uses FrameNet classes, it can directly be linked to this widely used resource in a multilingual setting. Our goal is to widen the resource in terms of coverage, by both integrating the existing event classes and by taking into account new lexical units and event classes from other domains. Given the considerable manual effort for populating such a resource, a semi-automatic strategy must be devised for this purpose.\smallskip
\item \textit{Anaphoric expressions}. As showed in the evaluation (especially in Section 6.3), the presented system is very sensitive to the lack of information in the structure of the given lexical unit. On the other hand, in discourse such information is often ``hidden'' due to anaphoric use of language (which is rather pervasive in Italian). Therefore, being able to resolve anaphoric expressions is a key step in order to improve both recall and precision of the system, without changing the overall methodology.\smallskip
\item \textit{Event Coreference}. \noindent The described graph architecture provides a starting point for further work on Event Coreference, that is, the task of linking different (textual) event mentions to the corresponding (real) event occurrence. Figure 5 is a suitable example, as it represents nine event mentions which are actually referencing to three real event occurrences: the injuring, arrest/capture and assassination of Duccio Galimberti. To this effect, our extraction pipeline could be used, alone or combined with other inputs, for producing a set of candidates for the coreference task. Being able to resolve automatically coreference between event mentions is our next step, as it would provide a very powerful tool for linking different information sources and to discover new information.  
\end{enumerate}

\appendix
\section{List of digitized texts}
This section contains the bibliographical entries of the digitized books that constitute the Memoirs and the Memoirs-test corpora (see section 3.1). Beside each entry, the corresponding abbreviation is specified. In square brackets we report the year of first publication, if it differs from the year of the edition used for OCR/digitization. Entries marked with a star (*) have been obtained by automatic text conversion from an existing digital edition.\smallskip

Memoirs corpus:
\begin{description}
\item[] Aimo Renato, \textit{Resistenza senza miti: dalla Stura alla Vesubie la brigata G.L. ``Carlo Rosselli''}, Cuneo, L'Arciere, 1991.
\item[] Artom Emanuele, \textit{Diari di un partigiano ebreo: gennaio 1940 - febbraio 1944}, a cura di Guri Schwarz, Torino, Bollati Boringhieri, 2008 [1966].
\item[] Balbo Adriano, \textit{Quando inglesi arrivare noi tutti morti. Cronache di lotta partigiana: Langhe 1943-1945}, Torino, Blu, 2005.
\item[] Berga Ugo, \textit{Diario partigiano: dall'8 settembre 1943 alla liberazione, gli eventi e le persone che coinvolsero la 106. brigata Garibaldi Giordano Velino}, s.l., s.n., stampa 2003 (Almese : Morra).
\item[] Bianco Alberto, \textit{Alberto Bianco: testimonianza partigiana}, a cura di Michele Calandri, Alessandra Demichelis, Savigliano, L'artistica, 1999.
\item[*] Bocca Giorgio, \textit{Partigiani della montagna: vita delle divisioni Giustizia e libert\`a del Cuneese}, Milano Feltrinelli, 2005 [1945].
\item[] Burdino Felice Luigi, \textit{Diario partigiano}, Pinerolo, Alzani, 2005.
\item[] Chiodi Pietro, \textit{Banditi}, Torino, Einaudi, 1975 [1946].
\item[] Comollo Gustavo, \textit{Il commissario Pietro, ANPI Piemonte}, stampa 1979 (Savigliano : Nuove Arti Grafiche).
\item[] Cordero Italo, \textit{Ribelle: esperienze di vita partigiana dalla Val Casotto alle Langhe dalla Liguria alle colline torinesi}, Mondov\`i, Fracchia, 1991.
\item[] Diena Marisa, \textit{Guerriglia e autogoverno: Brigate Garibaldi nel Piemonte occidentale 1943-1945}, Parma, Guanda, 1970.
\item[] Diena Marisa, \textit{Un intenso impegno civile: ricordi autobiografici del Novecento}, Torino, Lupieri, Fondazione Istituto piemontese Antonio Gramsci, 2006.
\item[] Donadei Mario, \textit{Cronache partigiane: la banda di Valle Pesio}, Cuneo, L'Arciere, 1980 [1973].
\item[] Dunchi Nardo, \textit{Memorie partigiane}, Cuneo, L'arciere, 1982.
\item[] Giaccardi Giovenale, \textit{Le formazioni R nella lotta di liberazione}, Cuneo, L'arciere, 1993 [1980].
\item[] Giovana Mario, \textit{Storia di una formazione partigiana: Resistenza nel Cuneese}, Torino, Einaudi, 1964.
\item[*] Marchesini Gobetti Ada, \textit{Diario partigiano}, Torino, Einaudi, 2014 [1956].
\item[] Martini Mauri Enrico, \textit{Partigiani penne nere: Boves - Val Maudagna - Val Casotto - Le Langhe}, Torino, Edizioni del Capricorno, 2016 [1968].
\item[] Modica Vincenzo, \textit{Dalla Sicilia al Piemonte: storia di un comandante partigiano}, 2. ed. riv. e corr, Milano, FrancoAngeli, 2003.
\item[*] Revelli Nuto, \textit{Le due guerre: guerra fascista e guerra partigiana}, Torino, Einaudi, 2005 [2003].
\item[*] Revelli Nuto, \textit{La guerra dei poveri}, Torino, Einaudi, 2014 [1962].
\item[*] Ronchi Della Rocca Icilio, \textit{Ricordi di un partigiano: la Resistenza nel Braidese}, a cura di Livio Berardo, Milano, Angeli, 2009 [1965].
\item[] Sacchetti Aldo, \textit{Un romano tra i ribelli: da Duccio Galimberti a Piero Cosa}, Cuneo, L'Arciere, 1990.
\item[] Trabucco Angela, \textit{Partigiani in Val Chisone: 1943-1945}, Torre Pellice, Tip. subalpina, 1959.
\item[] Visetti Aurelio Dante, \textit{Un ribelle come tanti: intorno ad un diario partigiano}, Cuneo, L'arciere, 1993.
\end{description}
Memoirs-test corpus:
\begin{description}
\item[] Aimo Renato, \textit{Il prezzo della pace: la gente bovesana e la Resistenza}, 1943-45, Cuneo, L'arciere, 1989.
\item[] Carminati Masera Diana, \textit{Langa partigiana '43-'45}, Boves, Araba Fenice, 2007 [1971].

\item[] Bianco Dante Livio, \textit{Guerra partigiana}, Torino, Einaudi, 2006 [1954].
\end{description}

\begin{acks}
The authors would like to thank Senan Kiryakos and Stella Peyronel for proofreading the article.
\end{acks}
\bibliographystyle{ACM-Reference-Format}
\bibliography{paper}

\end{document}